\documentclass[acmtog, authorversion]{acmart}

\AtBeginDocument{%
  \providecommand\BibTeX{{%
    \normalfont B\kern-0.5em{\scshape i\kern-0.25em b}\kern-0.8em\TeX}}}

\usepackage{graphicx}
\usepackage{caption}
\usepackage{subcaption}
\usepackage{array}
\usepackage{booktabs}
\usepackage{multirow}
\usepackage{xcolor}

\long\def\comment#1{}

\begin{document}

\title{Drawing Inductor Layout with a Reinforcement Learning Agent: Method and Application for VCO Inductors}


    \author{Cameron Haigh}
    \affiliation{%
      \institution{Huawei Noah's Ark Lab}
      \country{Canada}
    }
    
    \author{Zichen Zhang}
    \authornote{Both authors contributed equally to this research.}
    \author{Negar Hassanpour}
    \authornotemark[1]
    \affiliation{%
      \institution{Huawei Noah's Ark Lab}
      \country{Canada}
    }
    
    \author{Khurram Javed}
    \affiliation{%
      \institution{Huawei Noah's Ark Lab}
      \country{Canada}
    }
    
    \author{Yingying Fu}
    \affiliation{%
      \institution{Huawei Noah's Ark Lab}
      \country{Canada}
    }
    
    \author{Shayan Shahramian}
    \affiliation{%
      \institution{Huawei Noah's Ark Lab}
      \country{Canada}
    }
    
    \author{Shawn Zhang}
    \affiliation{%
      \institution{Huawei Noah's Ark Lab}
      \country{Canada}
    }
    
    \author{Jun Luo}
    \authornote{Corresponding author: jun.luo1@huawei.com}
    \affiliation{%
      \institution{Huawei Noah's Ark Lab}
      \country{Canada}
    }
\renewcommand{\shortauthors}
 {Haigh, et al.}

\begin{abstract}
Design of Voltage-Controlled Oscillator (VCO) inductors is a laborious and time-consuming task that is conventionally done manually by human experts.
In this paper, we propose a framework for automating the design of VCO inductors, using Reinforcement Learning (RL).
We formulate the problem as a sequential procedure, where wire segments are drawn one after another, until a complete inductor is created.
We then employ an RL agent to learn to draw inductors that meet certain target specifications.
In light of the need to tweak the target specifications throughout the circuit design cycle, 
we also develop a variant in which the agent can learn to quickly adapt to draw new inductors for moderately different target specifications.
Our empirical results show that the proposed framework is successful at automatically generating VCO inductors that meet or exceed the target specification.
\end{abstract}




\maketitle

\section{Introduction}
Voltage Controlled Oscillators (VCOs) are circuits that generate an oscillating signal whose frequency is controlled by an input voltage. 
As such, VCOs are widely used in Radio Frequency (RF) applications where there is a need to tune the device within a certain range of frequencies.
A common form of VCO, 
which we focus on in this paper,
is the Inductance Capacitance VCO that uses the coupling between an inductor and a capacitor to generate the oscillating signal.
In this case, the capacitor is tuned by a voltage input to modify the frequency.
To achieve a wide tuning range with a VCO,
the effective resistance of the inductor should remain low, 
the Self Resonance Frequency (SRF) should be high, 
and most importantly, the inductance should stay close to a target value.

Since the shape%
\footnote{Also referred to as ``layout'' in this paper.}
of an inductor determines its effective inductance and resistance,
it is important to design it in such a way that the resulting inductor meets the \textit{performance} requirements 
(i.e., desired specifications such as the effective inductance, resistance, etc.).
There are, however, clear and strict \textit{design} requirements 
(e.g., maximum available area, no crossing wires, wire turn angle constraints, etc.) 
that are imposed by manufacturing considerations.

\begin{figure}[t]
    \centering
    \includegraphics[width=0.45\columnwidth]{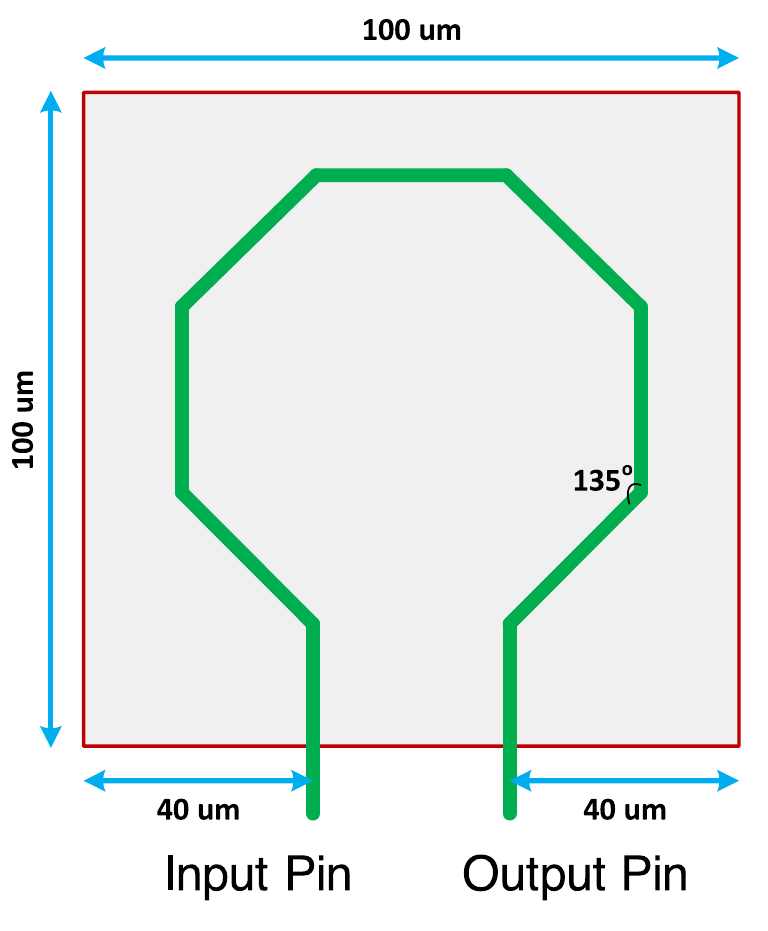}
    \caption{Canvas properties of an example inductor layout.}
    \label{fig:canvas}
\end{figure}

Conventionally, the layout of an inductor is designed manually by human experts (see for example, Figure~\ref{fig:canvas}); 
however, this is a complex and time-consuming task.
It involves an iterative process with two major steps: 
{
a human expert first designs/modifies an inductor layout to meet a certain target specification, 
and analyzes the simulation performance of the inductor in a larger circuit\footnote{
{Designers often use a template design (e.g., an octagon) and tune its parameters such as wire length and width to achieve their target specification.}
}
.}
In this work, we formulate the task as an optimization problem that takes into account the above-mentioned design and performance requirements while generating inductor layouts.
{
Our algorithm can come up with a variety of layout shapes, all of which can meet or exceed%
\footnote{
{
I.e., coming up with a design that for example takes less area or has a larger SRF, etc.
}
}
the target specifications. 
This is beneficial for cases where the canvas shape is irregular for example. 
Moreover, some of produced layouts might take a smaller area than what the human designers have come up with, which is desirable.
}

We formulate the design framework for VCO inductors as a \textit{drawing task}.
Specifically, it is formulated as a Markov Decision Process (MDP)~\cite{puterman2014markov},
where an agent draws a single wire segment on the canvas at each time step.
Hence the framework is much more flexible compared to the template-based methods
(elaborated in the Related Works section) 
in terms of the variety of shapes it can generate.

Moreover, our framework allows for \textit{several optimization techniques}%
\footnote{
E.g., Genetic Algorithm (GA), Bayesian Optimization (BO), Reinforcement Learning (RL), etc.
}
to be used to generate candidate layouts.
Here, we focus on using Reinforcement Learning (RL) \citep{sutton2018reinforcement}.
We show that the RL agent is able to generate candidate designs with performance matching the target specifications. 
We also develop a variant that \textit{adapts faster} than the alternative optimization methods when the target performance requirements are [mildly] shifted.%
\footnote{
    This is a common scenario in practice; where the human experts tweak the target specifications frequently to achieve the best performance for the larger circuit.
}

\section{Related Works}

\begin{table*}[t]
\centering
\setlength{\tabcolsep}{6pt} 
\renewcommand{\arraystretch}{1.25} 
\begin{tabular}{l p{28mm} p{15mm} p{30mm} p{30mm}}
\toprule
	Methods & Reliance on \newline Expert Knowledge & Sample \newline Efficiency	&Design \newline Space & Transfer to \newline Different Targets	\\
\midrule
Template Optimization &	High	&	High 	&	Highly Constrained & Hard	\\
Evolved Antenna &	? &	? &	Less Constrained & Hard \\
Our Proposed Solution & Low & High&	Less Constrained & Easy 	\\
\bottomrule
\end{tabular}
\caption{Comparison of the proposed method and the existing approaches.}
\label{tab:comparison}
\end{table*}

{
Note that the goal for VCO inductor design is to find the best geometry that can even \emph{exceed} the target specifications. 
This goal is different from the majority of literature on optimizing passive designs 
since they are often focused on only \emph{meeting} the target specifications with the minimum number of simulations. 
VCO inductors are special because there are usually only a few of them on a chip, 
so finding the optimal one is quite important.
}

Existing methods related to the automated layout design for VCO inductors are summarized in Table~\ref{tab:comparison}, 
where the characteristics of each method is compared to the proposed approach.
Description of each existing method follows.

\subsection{Template Optimization}
Current Electronic Design Automation (EDA) tools for inductor design formulate their solution as optimizing only a certain set of parameters of a fixed template layout.
That is, the shape of the layout is pre-determined (e.g., a spiral),
and the optimization procedure searches through the valid values of the layout’s parameters (e.g., number of turns, wire width, wire spacing, etc.) 
in order to find inductors that meet the target specification. 
Various optimization approaches have been used in the literature: 
Genetic Algorithm (GA) \citep{farhat2015optimization, liu2011synthesis}, 
Heuristic Techniques \citep{hajjami2020shape}, 
Bayesian Optimization (BO) \citep{torun2018global},
{
Evolutionary Computation and Gaussian-Process Surrogate Modeling \citep{passos2017radio}, and 
Corner-Aware Optimization \citep{passos2018enhanced}.
}

The problem with the template optimization approach is two-fold:
(i)~the rigidity of the pre-determined layout that hugely limits the search space for valid inductors, 
which in turn results in designing sub-optimal layouts; and 
(ii)~inability to quickly come up with a new design that meets a moderately different target specification
(i.e., the entire algorithm must be re-run and no knowledge transfer is possible from previously learned models).

\subsection{Evolved Antenna}
\citet{hornby2006automated} converted the problem of antenna design into a 3D drawing problem. 
Their proposed algorithm applies GA to open-ended sequences of line segments, 
where the mutations include:
(i)~adding to the length of each line segment; and 
(ii)~rotating each line segment along x, y, and z axes.

To the best of our knowledge, such a step-by-step method that involves drawing has not been used to tackle VCO inductor design;
and it is not trivial how to convert the antenna design to a VCO inductor design problem either.
This is because the former requires evaluation of an incomplete antenna at each time step while the VCO inductor 
{
does not require that. The inductor 
}
must be completed 
(i.e., the input port is wired to the output port) 
before its performance is evaluated through simulations
{
(this is beneficial since queries to the simulator are quite time-consuming).
}
Moreover, similar to the template optimization methods, 
this approach is not able to quickly come up with a new design 
that meets a moderately different altered performance requirements than the original one.

\section{VCO Inductor Design as a Drawing Task}
In our formulation of the VCO inductor design process, an inductor is described by a sequence of segments,
where at each step, knowledge of the past steps is used to determine the placement of a new segment.
Each segment contains information on its wire length, width, angle relative to the previous segment, 
and the metal layers on the chip which will contain this segment. 
This information is sufficient in order to manufacture a designed layout as an on-chip inductor.
The first segment is always initiated at the input port
(the location is determined by 
{some parameters of } 
the circuit), 
and each subsequent segments are relative to the one before them. 
A design is considered invalid if wire collision has happened or some portion of the inductor extends outside the maximum available area.
Designs are considered complete when a segment has reached the output port for non-symmetric designs or the vertical mid-line for symmetric designs.

The action space corresponds to the information required for placing each segment. 
To constrain the huge search space and keep the experiments simple, 
the designs were enforced to have a fixed wire width across all segments
\footnote{
    Although the wire width could be added to the action space (as a dimension), 
    for our setting, we found the respective improvement to be empirically insignificant.
}
and the length of each segment was fixed. 
As such, the action space was reduced to covering only the angle of the segment relative to the previous one. 
To comply with the design requirements, the angles are discretized, with actions {0, 1, 2, 3, and 4} 
corresponding to the angles of {$-90^{\circ}$, $-45^{\circ}$, $0^{\circ}$, $45^{\circ}$, and $90^{\circ}$} respectively.%
\footnote{
{
Note that this is a manufacturing constraint that we embedded into our method. 
Our formulation is general and the actions can use finer discretizations of the angles based on the need and requirements.
}
}

As such, our method is quite flexible with the layouts that it can generate.
Thus, we can design a valid inductor even when our available space is not evenly shaped (e.g., a perfectly square canvas) with a non-symmetric design.
Conventionally, however, symmetric designs are preferred in practice to preserve symmetry in the overall circuit. 
Hence in our paper, we focus on symmetric cases without loss of generality. 
In this case, the drawing task terminates when it reaches the vertical mid-line between the input and output ports, 
and the segments are mirrored to complete the inductor.

\begin{figure*}[t]
    \centering
    \includegraphics[width=0.7\textwidth]{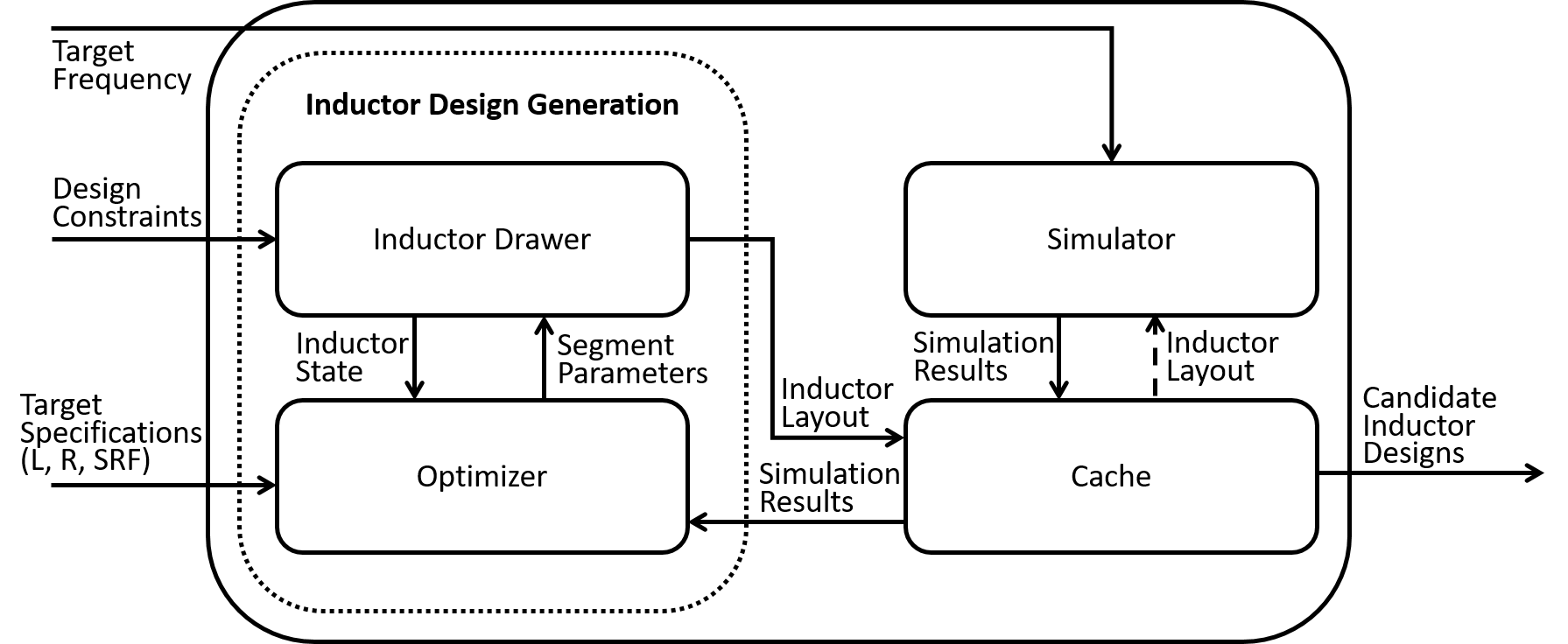}
    \caption{The proposed process for drawing a VCO inductor.}
    \label{fig:modules}
\end{figure*}

This formulation can be applied to arbitrary sequential optimization techniques by
(i)~providing the technique with some representation of the current segments, and 
(ii)~receiving from the technique the parameters for drawing the next segment. 
For non-sequential baseline methods that operate on a fixed set of parameters, 
a fixed number of segments can be used as the optimization target, 
considering only as many segments from the result as are needed to create an inductor design and discarding the rest.

To summarize, the proposed framework is comprised of the following four modules 
(see also Figure~\ref{fig:modules}), 
which all interact with each other to produce the overall solution:
\begin{enumerate}
    \item The \textbf{inductor drawer} module is in charge of producing the inductor designs.
    It creates an inductor by adding line segments
    (each describing a segment of wire that can be realized)
    to the canvas, starting from a specified input position and appending each segment onto the previous one. 
    These segments are each drawn using several parameters provided by the optimizer module.
    
    \item The \textbf{optimizer} module is in charge of ensuring that the performance of the produced inductors will be improving with respect to their target specifications. 
    Feedback is given to the optimizer in terms of how well designs met the target specifications or how badly they violated the design constraints.
    This feedback allows the optimizer to improve its model, such that it suggests parameters that result in designing more optimal layouts.
    
    \item The \textbf{cache} module records all valid inductor designs that have been generated as well as their  evaluated performance.
    The cache is used for fast retrieval of the performance measures of already seen layouts so that the number of queries to the [slow] simulator is minimized.
    
    \item The \textbf{simulator} module takes the newly produced inductor designs from cache%
    \footnote{
    {
    This is done indirectly from the cache: 
    Once a new inductor is drawn, it only gets simulated if it is not already in the cache, 
    in which case it is first recorded in the cache. 
    The simulator then reads the drawn inductor from the cache as input (the dashed line on the figure) and evaluates it.
    }
    }
    , evaluates them, 
    and returns their performance specifications --- i.e., L (inductance), R (resistance), SRF, Q (quality factor), etc. to be recorder in the cache.
    Processing each query to the simulator is often the bottle-neck in the procedure.
\end{enumerate}

\section{Reinforcement Learning for Optimization}
We use RL to solve the drawing task by serving as the optimizer module of the proposed framework. 
For that, we formulate the optimizer module's task as an MDP, denoted by a tuple $(S, A, R,  P, \rho, \gamma)$, 
where
{
$S$ and $A$ are the finite state space and finite action space respectively;}
$R:~S~\times~A~\rightarrow~\mathcal{R}$ is the reward function; 
$P(s,a,s')$ is the transition probabilities%
\footnote{
{
In our task, the transition function is deterministic.
} 
from state $s$ to state $s'$ when taking action $a$
};
$\rho$ is the state distribution of the initial state $s_0$ which is deterministic in our task (i.e., the first wire segment always starts at the input port); 
and
$\gamma~\in~[0,1]$ is the discount factor.

The goal of the design task is to maximize the expected sum of discounted rewards (i.e., the return) 
$J_\pi~=~\mathbb{E}_\pi~\Big[\sum_{t=0}^T~\gamma^t R(s_t,a_t)\Big]$ 
where $s_t$ and $a_t=\pi(s_t)$ are the state and action at time step $t$ and 
policy $\pi(\cdot)$ is a function that maps a state to an action. 
$T$ is the termination time step of the episode. 
We set discount factor $\gamma=1$ since this is an episodic problem with sparse rewards 
where the reward is non-zero only at terminal states and zero elsewhere.%
\footnote{
{
The problem of sparse rewards is not prominent in this application since our episode lengths are rather short (i.e., approximately less than 15 steps each)
}
}

In our formulation, an episode corresponds to one attempt at drawing an inductor, 
whether the resulting inductor is valid or not.
A valid inductor is one whose input port is successfully connected to its output port, without any wire collisions.
In order to maximize efficiency, we added a masking mechanism in which the action(s) that result in colliding to another wire are eliminated from the agent's search space. 
This helps preserving what the agent has drawn so far and attempt to learn on top of that.


To ensure that the inductors are being optimized for the performance requirements, we design our reward signal accordingly.
The reward is zero for non-terminal states 
(i.e., no feedback is given to an incomplete inductor). 
As soon as an episode is terminated
(i.e., either when wires collision or area perimeter collision is inevitable; 
or the input and output ports are successfully connected)
a reward is given. 
If the former was true, i.e., the generated layout was an invalid design, a small penalty is given;
otherwise the reward is determined from how well the designed inductor's specifications meets the performance requirements.
The \textbf{reward} function for a complete inductor is defined as:
\begin{equation}
    1 - 
    \frac{\omega_{\scriptscriptstyle L} \!\times\! C_{\scriptscriptstyle L} +\omega_{\scriptscriptstyle R} \!\times\!  C_{\scriptscriptstyle R} + \omega_{\scriptscriptstyle SRF} \!\times\! C_{\scriptscriptstyle SRF} + \omega_{\scriptscriptstyle Area} \!\times\! C_{\scriptscriptstyle Area}}
    {\omega_{\scriptscriptstyle L}+\omega_{\scriptscriptstyle R}+\omega_{\scriptscriptstyle SRF}+\omega_{\scriptscriptstyle Area}}
\end{equation}
where $C_{\scriptscriptstyle L}$, $C_{\scriptscriptstyle R}$, $C_{\scriptscriptstyle SRF}$, and $C_{\scriptscriptstyle Area}$ 
stand for the cost of inductance, resistance, SRF, and area respectively; and
$\omega_{\scriptscriptstyle L}$, $\omega_{\scriptscriptstyle R}$, $\omega_{\scriptscriptstyle SRF}$, and $\omega_{\scriptscriptstyle Area}$ 
stand for the weights of each respective cost.

\comment{

{\begin{equation*}
1 - \frac{w^{\top}C}{\left\Vert w \rightarrow\Vert_{1}}
\end{equation*}
 where w is the weight vector [$w_L$, $w_R$, ...] and C is the cost vector [$C_L$, $C_R$ ...]}
}
The cost values are computed as follows:

\begin{align*}
C_{\scriptscriptstyle L} & = \begin{cases}
E_{\scriptscriptstyle L} & E_{\scriptscriptstyle L} < 0.05 \\
2E_{\scriptscriptstyle L} - 0.05 & E_{\scriptscriptstyle L} \geq 0.05
\end{cases} \\
C_R & = \begin{cases}
0 & E_{\scriptscriptstyle R} < 0 \And E_{\scriptscriptstyle L} \geq 0.05 \\
E_{\scriptscriptstyle R} & E_{\scriptscriptstyle R} < 0 \And E_{\scriptscriptstyle L} < 0.05  \\
min(2E_{\scriptscriptstyle R}, 1) & E_{\scriptscriptstyle R} \geq 0
\end{cases} \\
C_{SRF} & = \begin{cases}
0 &  E_{\scriptscriptstyle SRF} < 0  \And E_{\scriptscriptstyle L} \geq 0.05 \\
max(2E_{\scriptscriptstyle SRF}, -1) & E_{\scriptscriptstyle SRF} < 0 \And E_{\scriptscriptstyle L} < 0.05  \\
min(2E_{\scriptscriptstyle SRF}, 1) & E_{\scriptscriptstyle SRF} \geq 0
\end{cases} \\
C_{Area} & = \begin{cases}
0 & E_{\scriptscriptstyle L} \geq 0.05 \\
E_{\scriptscriptstyle Area} & E_{\scriptscriptstyle L} < 0.05 \\
\end{cases}
\end{align*}
where
$E_{\scriptscriptstyle L} = \Big|\frac{L}{L_T}-1\Big|$,
$E_{\scriptscriptstyle R} = \frac{R}{R_T}-1$,
$E_{\scriptscriptstyle SRF} = 1 - \frac{SRF}{SRF_T}$,
$E_{\scriptscriptstyle Area} = \frac{\text{Area}}{\text{Area}_{\scriptscriptstyle \text{MAX}}} - 1$,
and $L_T$, $R_T$, $SRF_T$ are the target specifications and $\text{Area}_{\scriptscriptstyle \text{MAX}}$ is the maximum area available for the canvas.

{
The reward function is a linear combination of the cost for each performance requirement 
(i.e., inductance, resistance, SRF, and area). 
Each individual cost component as shown above is a piece-wise linear function of the error 
(`$E$' terms in the equations) 
for the corresponding requirement that is designed in consultation with the human experts. 
For example, the experts mentioned that $5\%$ is an acceptable range for error in inductance. 
That is the reason for 0.05 being the breaking point in the sub-domains in the respective piece-wise function for the inductance cost. 
The penalty increases faster when the inductance is outside the $5\%$ acceptable range.
Similar rules are applied to the rest of parameters.
}

Due to the inherently spatial nature of the design problem, 
the state of the drawing is represented as a Boolean image where pixels are true for locations that are covered by an inductor segment and false for locations that are not. 
We train a Deep Q-Network (DQN)~\citep{mnih2015human} agent to solve this task. 
DQN takes advantage of Deep Learning (DL)~\citep{goodfellow2016deep}, 
specifically Convolutional Neural network (CNN) architecture~\citep{lecun1989backpropagation}, 
in order to learn useful feature representations from the raw image states.
DQN follows a conventional algorithm in RL, namely Q-learning~\citep{watkins1989learning}, 
to learn the action-values from those learned representations.
The action-value function is then used to learn optimal policies for drawing inductors that meet the target specifications.

\subsection{Transfer Learning}
Once trained, the RL agent described in the previous section can draw inductor layouts that meet a certain, pre-determined target specification.
However, such an ad hoc agent is not capable of producing layouts that meet new, moderately tweaked targets. 
However, due to circuit limitations uncovered throughout the circuit design cycle, the targets need to be tweaked throughout.
To address this necessity, we modified the implementation of the RL agent so that it would take less time to produce new candidate VCO inductors when the target is changed during the design process.
We do so by expanding the state observation such that it includes the desired target specifications (L, R, and SRF) as input.
The respective neural network architecture for these features is a fully connected network.

To facilitate transfer learning for the RL agent,
we divide the training into two stages. 
First, we pre-train the agent on an initial \textit{distribution} of targets around the reference target.
The distribution is centered on the reference target and only covers a small range around it 
(e.g., 20\% above and below the reference target). 
An agent is trained on targets sampled from this distribution to the full.
Then, once a new target is introduced, we fine-tune the trained agent for this new target.
We empirically show that fine-tuning an agent that is pre-trained on different (yet close) targets is much faster than training a separate agent from scratch for each new target --- see the Results and Discussions section.

\section{Experiments}
\subsection{Setup}
In our experiments, we used an area of $100 \times 100$ $\mu$m (i.e., canvas)
with the input and output ports located at the bottom of the allotted area at $40$ $\mu$m and $60$ $\mu$m respectively (see Figure~\ref{fig:canvas}).
We fixed the length of the segments that can be drawn such that the ends of each segment land on a $10 \times 10$ $\mu$m grid 
(i.e., steps in the four cardinal directions had a length of $10$ $\mu$m and diagonal steps had a length of $\sqrt{2} \times 10$ $\mu$m). 
We also fixed the width of all segments to $5$ $\mu$m and kept all segments on a single metal layer. 

Figure~\ref{fig:canvas} illustrates a commonly used expert design for a VCO inductor following the above-mentioned restrictions. 
With slightly relaxed requirements on resistance and SRF,
this design yields a desired inductor performance of $L~=~116.5$~pH, $R~\leq~0.925$~$\Omega$, and $SRF~\geq~155$~GHz, at 15 GHz operating frequency. 
To determine the performance of the generated designs, we used ASITIC (Analysis and Simulation of Spiral Inductors and Transformers for ICs) \citep{asitic}, 
which is an open source simulator for on-chip RF components.
Cadence\textsuperscript{\textregistered} Virtuoso\textsuperscript{\textregistered} \cite{CadenceVirtuoso} and EMX \cite{EMX, emx_paper} were used for layout and simulation of chosen inductors as an additional validation step to ensure consistent results. 

\subsection{Results and Discussions}
\subsubsection{Symmetry}
\begin{figure}[t]
    \centering
    \includegraphics[width=0.75\columnwidth]{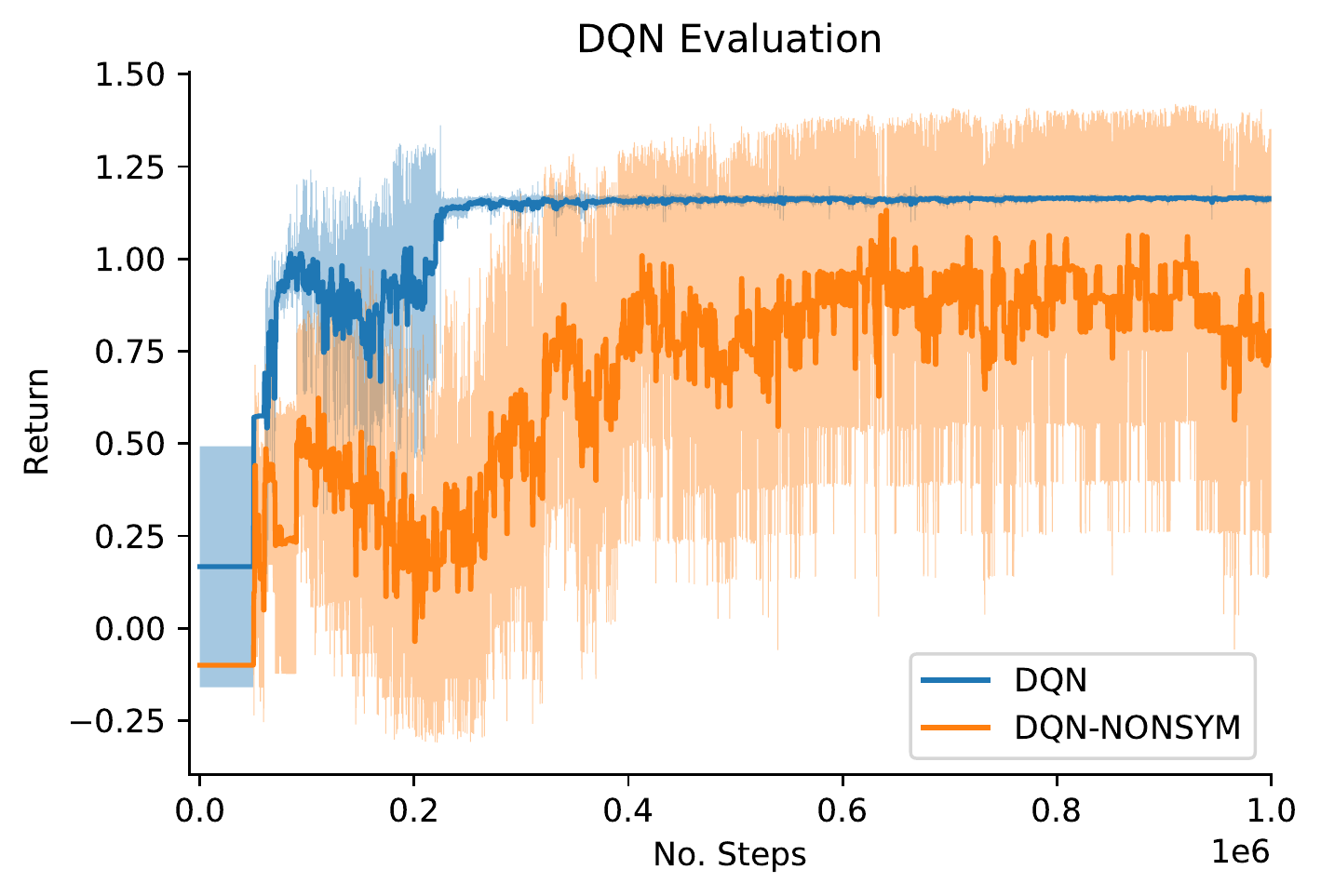}
    \caption{Performance of the inductor generated by the trained model at different time steps in training. The line and shaded region correspond to the mean and standard deviation from 15 independent runs.}
    \label{fig:DQN_eval}
\end{figure}

\begin{figure}[t]
    \centering
     \begin{subfigure}{0.99\columnwidth}
    \includegraphics[width=\linewidth]{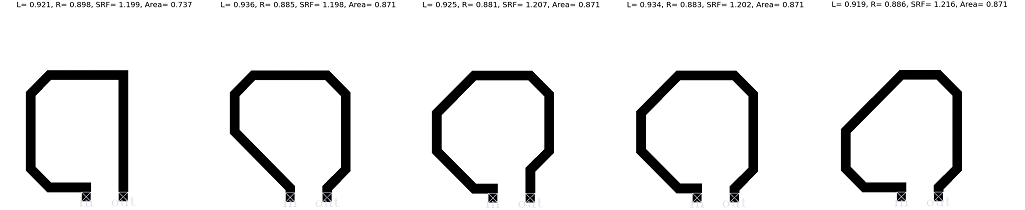}
    \caption{DQN, non-symmetric}
    \label{subfig:b}
    \end{subfigure}   
    \begin{subfigure}{0.99\columnwidth}
    \includegraphics[width=\linewidth]{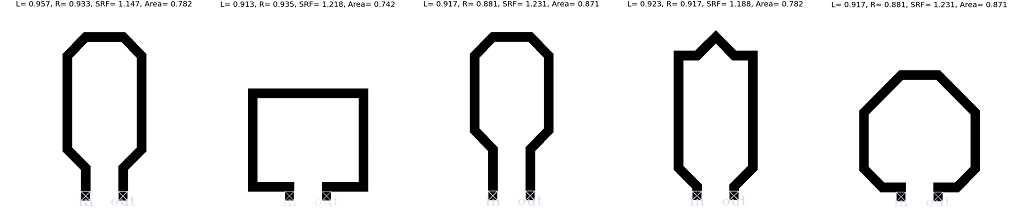}
    \caption{DQN, symmetric}
    \label{subfig:c}
    \end{subfigure}
    \begin{subfigure}{0.99\columnwidth}
    \includegraphics[width=\linewidth]{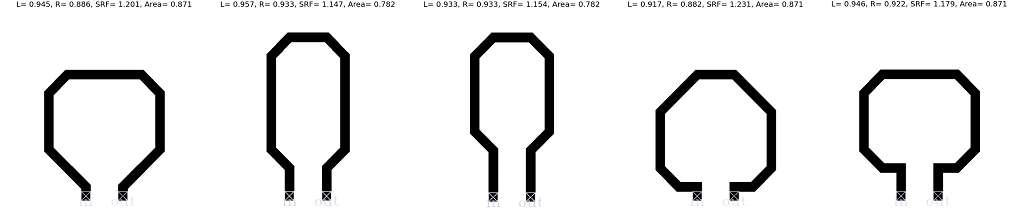}
    \caption{GA, symmetric}
    \label{subfig:d}
    \end{subfigure}
    \begin{subfigure}{0.99\columnwidth}
    \includegraphics[width=\linewidth]{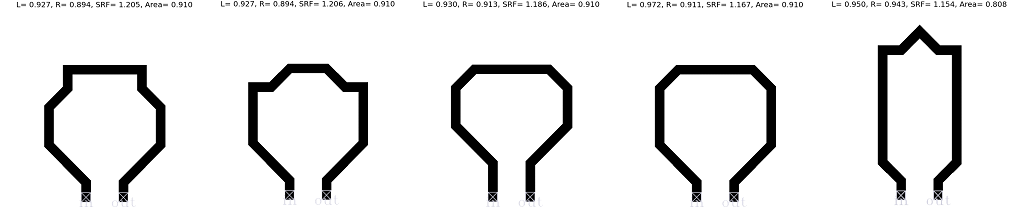}
    \caption{Random, symmetric}
    \label{subfig:e}
    \end{subfigure}
    \caption{Visualization of the top 5 inductors ranked by the return (sum of rewards), designed by the respective algorithms:
    (\subref{subfig:b})~non-symmetric DQN,
    (\subref{subfig:c})~DQN,
    (\subref{subfig:d})~GA, and
    (\subref{subfig:e})~Random Agent.
    Numbers on top of each layout are their respective simulated results from ASITIC, normalized by the target specification. 
    Better viewed when zoomed in. 
    The reference design in Figure~\ref{fig:canvas} has $L=1,R=0.974,SRF=1.02,\text{Area}=1$. 
    The results indicate that the algorithms can produce inductors exceeding the reference human design, 
    e.g., having a smaller area while achieving similar targets.}
    \label{fig:layouts}
\end{figure}

We chose primarily to focus on generating symmetric designs.
To validate the claim that we do not lose any performance,
we compared the training performance of a DQN agent producing symmetric designs with one using non-symmetric designs. 
From Figure~\ref{fig:DQN_eval}, we can see that while we get some very minor performance gain {in terms of the upper bound of return} from using non-symmetric designs, 
it comes at the cost of much longer training time as well as high variance. 
In contrast, when using symmetric designs, we converge to an optimal design with far fewer interactions with the environment.
Figures~\ref{fig:layouts}~(\subref{subfig:b})~and~(\subref{subfig:c}) illustrate the top 5 layouts for symmetric and non-symmetric settings respectively.

\subsubsection{Comparison to Baseline Methods}
In this work, we use DQN (RL agent) as the optimization module to design the inductors in a sequential manner. 
However, there is some precedent to use a one-shot method to produce inductors. 
Therefore, we compared the results of DQN to that of GA as the optimization module (see Figure~\ref{fig:modules}).
Using GA, we optimize a sequence of 15 segments at once to create the inductors, 
disregarding any trailing segments that come after the segment has reached the output port. 
The sequence length was chosen to be 15 since it is larger than the length of the top performing designed inductors,
giving the algorithm the opportunity to create designs as good as the ones that the RL agent could produce. 
{
We also compared our DQN agent with a baseline random agent that generated layouts in the same sequential manner as RL where it selected a random action at each step.}

\begin{figure}
    \centering
    \includegraphics[width=0.75\columnwidth]{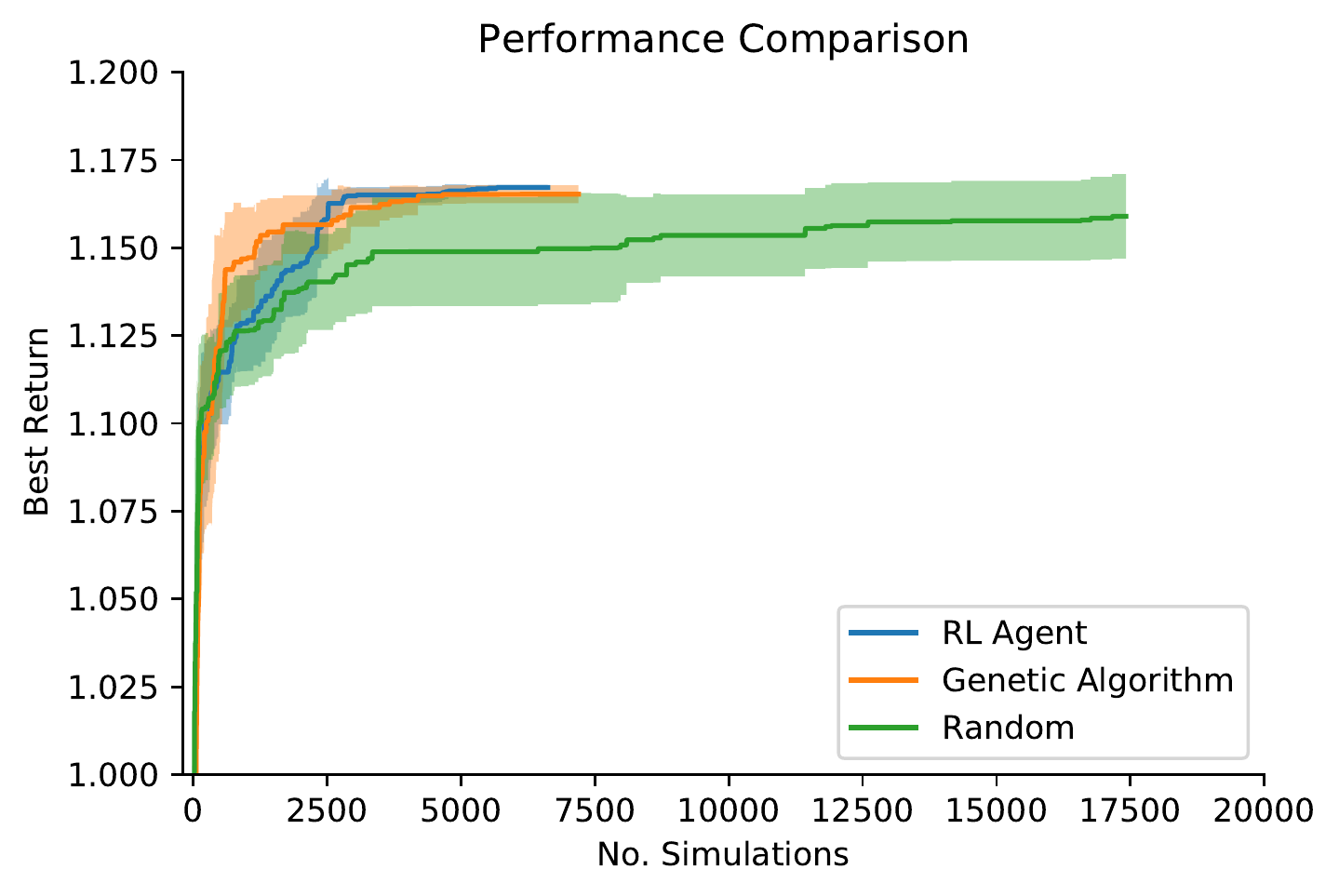}
    \vspace{-5pt}
    \caption{Best inductor performance during training with respect to number of simulated inductors.
    The line and shaded region correspond to the mean and standard deviation from 12 independent runs.}
    \label{fig:baselines}
\vspace{-5pt}
\end{figure}

{
From Figure~\ref{fig:baselines}, we can see that DQN and GA significantly outperform the random agent in terms of number of simulations, 
meaning that they require far fewer queries to the simulator to converge to a good agent that is able to draw inductors meeting the performance requirements.
Moreover, although GA achieves higher rewards at first, 
DQN surpasses GA at higher number of simulations.
We can also see that DQN exhibits less variance than GA.
Since we desire that our method to consistently produce the best inductor possible for a given specification, 
the DQN agent is preferred. 
Figures~\ref{fig:layouts}~(\subref{subfig:d})~and~(\subref{subfig:e}) illustrates the top 5 layouts generated by the GA and random agents respectively.
}

\begin{figure}
    \centering
    \includegraphics[width=0.75\columnwidth]{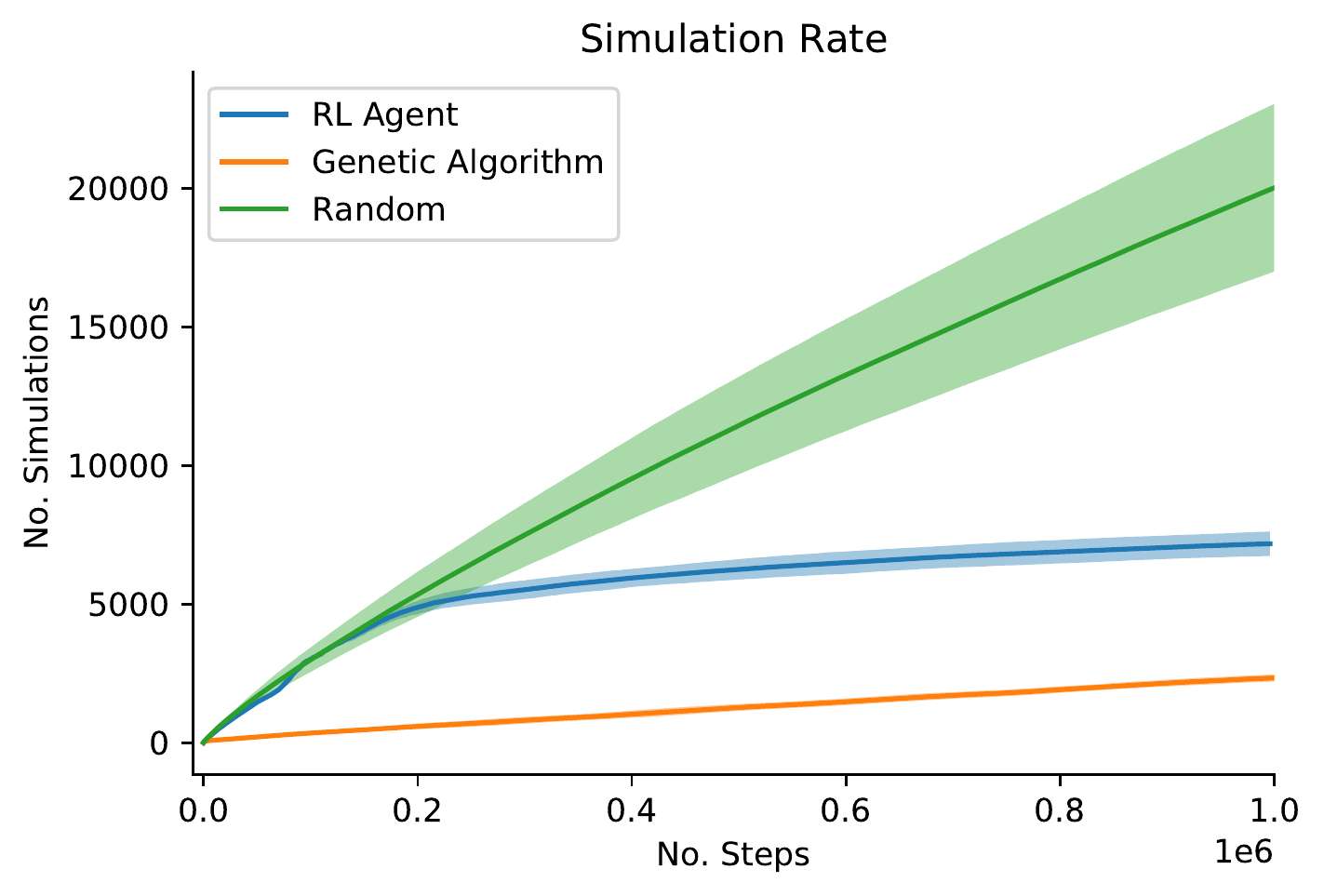}
    \vspace{-5pt}
    \caption{Growth of number of simulations versus number of steps.
    Higher slope indicates the corresponding method generates more \emph{new} layouts that are not in cache. 
    The line and shaded region correspond to the mean and standard deviation from 12 independent runs.
    }
    \vspace{-5pt}
    \label{fig:versus}
\end{figure}

Figure~\ref{fig:versus} illustrates the growth of number of simulations versus steps for all three agents.
Although the random agent keeps generating novel inductors%
\footnote{
Recall that we only send the novel inductors to the simulator for evaluation. 
We cache the evaluated inductors in order to reduce the time spent in the simulator bottle-neck.
}
as the number of steps increases,
it does not improve the achieved rewards as much as GA and DQN (Fig.~\ref{fig:baselines}).
The GA agent, on the other hand, produces many duplicate inductors,
{
failing to properly explore the space of valid inductors.
Part of the reason is that some mutations only happened in the trailing segments after the layout was finished which did not generate new inductors.
It explains its smaller mean and standard deviation compared to the other methods.}
The DQN agent, however, seems to have reached a good balance in terms of exploration of novel inductors and exploitation of the inductors that it has already generated.

\subsubsection{Transfer Learning}
\begin{figure}[t]
\centering
\includegraphics[width=0.75\columnwidth]{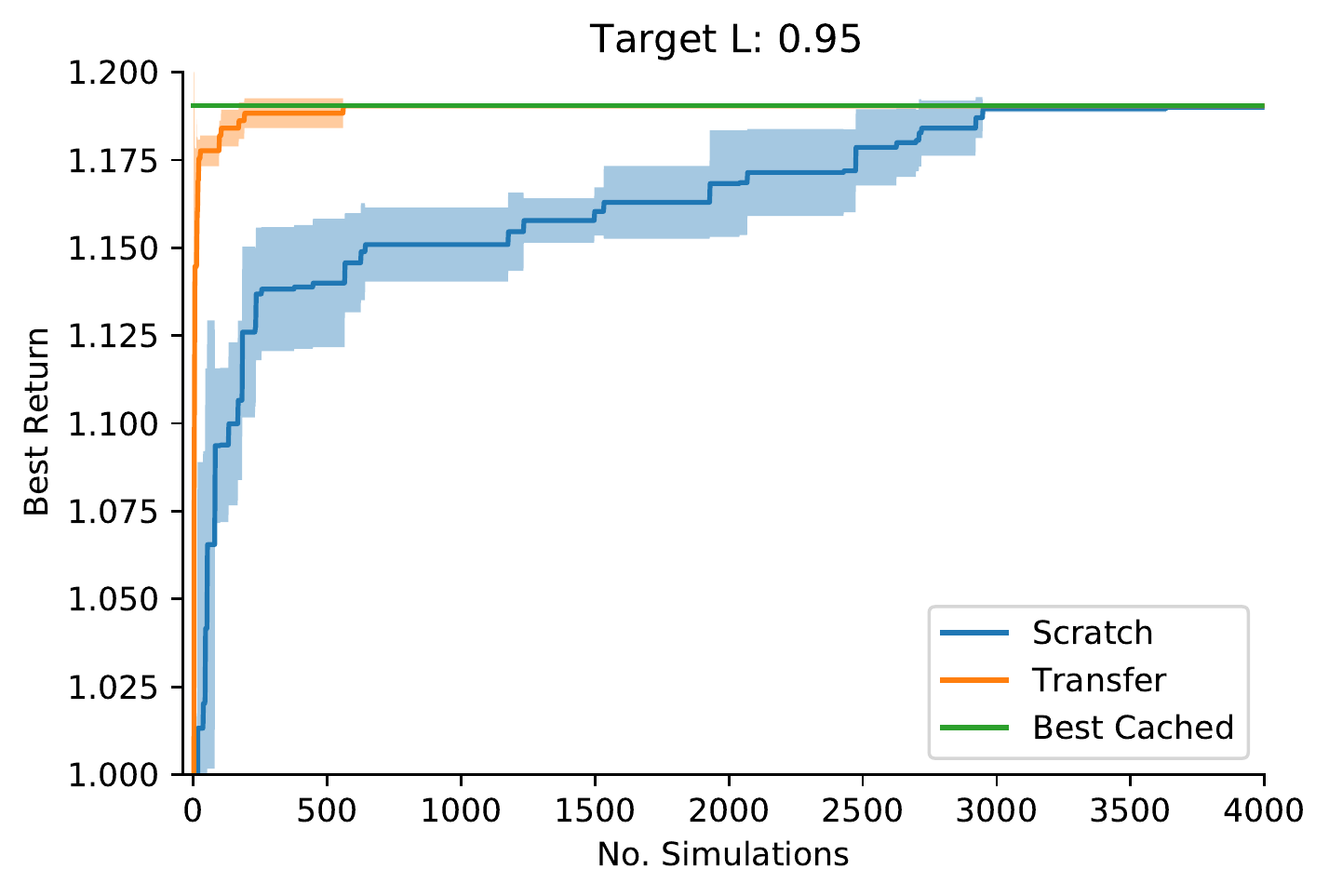}\\
    \vspace{.25cm}
\includegraphics[width=0.75\columnwidth]{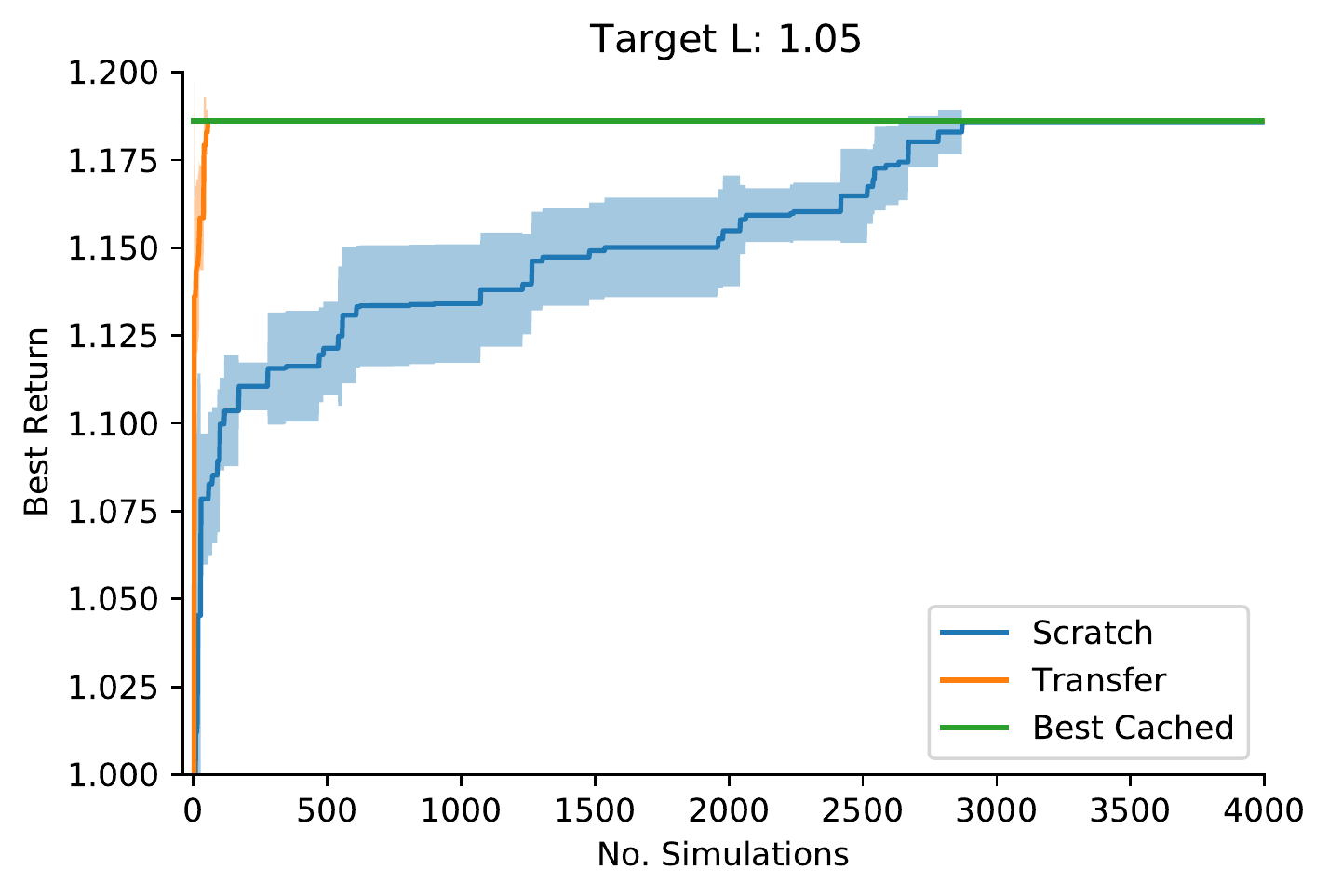}
\caption{Comparison of transferred model (orange) versus a model trained from scratch (blue) in terms of the number of simulations to achieve the best return. The line and shaded region correspond to the mean and standard deviation from 5 independent runs.
{
The ``Best Cached'' line (green) shows the best achieved reward ever, either by the transfer or scratch agent.}
}
\label{fig:transfer}
\vspace{-10pt}
\end{figure}

Here, the task is to generate inductors for a new target specification. 
There are two ways to do this: (i)~train from scratch; and (ii)~fine-tune a pre-trained model on similar targets (i.e., transfer).
{
To facilitate transfer learning for the RL agent, we divide the training into two stages. 
First, we pre-train the agent on an initial distribution of targets around the reference target. 
An agent is trained on targets sampled from this distribution to the full. 
Then, once a new target is introduced, we fine-tune the trained agent for this new target. 
Figure~\ref{fig:transfer} illustrates the performance for an agent that is trained from scratch (in blue) and transfer (in orange)
and we show the results for when the new target is set to L=0.95 (top) and L=1.05 (bottom). 
In both cases, we can see that training from scratch to achieve the new target takes much longer to generate a top performing layout than the transferred model.
}

\section{Conclusion}
We proposed and implemented an Electronic Design Automation (EDA) tool that automates the design of Voltage-Controlled Oscillator (VCO) inductors. 
This tool can
(i)~generate several valid VCO inductors designs 
(a valid design being one which obeys the design constraints and meets the target specifications); and 
(ii)~quickly produce new candidate inductors when the target specifications are moderately tweaked. 
We formulated the task as a drawing problem and solved it using Reinforcement Learning (RL) as the core optimization technique.
We empirically showed that our tool can match or exceed the performance of the simpler human designs. 

The current framework is focused solely on the design of VCO inductors, 
however, it could be more broadly applied to the design of general inductors 
{
(e.g., multi-layer architecture).
} 
This framework can also be applied to the design of more complex VCO inductors; 
for instance, to a switchable VCO inductor design involving multiple distinct coils,
each drawn in the same manner as we draw a single coil here.
These are left to future work.

\bibliographystyle{ACM-Reference-Format}
\bibliography{draft}
\end{document}